\documentclass{article}

\usepackage{times}
\usepackage{graphicx} 
\usepackage{subfigure} 

\usepackage{natbib}

\usepackage{algorithm}
\usepackage{algorithmic}

\usepackage{hyperref}

\usepackage[accepted]{icml2017}

\icmltitlerunning{A Lightweight Front-end Tool for Interactive Entity Population}

\begin{document} 

\twocolumn[
\icmltitle{A Lightweight Front-end Tool for Interactive Entity Population}



\icmlsetsymbol{equal}{*}

\begin{icmlauthorlist}
\icmlauthor{Hidekazu Oiwa}{equal,recruit}
\icmlauthor{Yoshihiko Suhara}{equal,recruit}
\icmlauthor{Jiyu Komiya}{recruit}
\icmlauthor{Andrei Lopatenko}{recruit}
\end{icmlauthorlist}

\icmlaffiliation{recruit}{Recruit Institute of Technology}

\icmlcorrespondingauthor{Hidekazu Oiwa}{oiwa@recruit.ai}
\icmlcorrespondingauthor{Yoshihiko Suhara}{suharay@recruit.ai}

\icmlkeywords{boring formatting information, machine learning, ICML}

\vskip 0.3in
]



\printAffiliationsAndNotice{\icmlEqualContribution} 

\begin{abstract}
Entity population, a task of collecting entities that belong to a particular category, has attracted attention from vertical domains.
There is still a high demand for creating entity dictionaries in vertical domains, which are not covered by existing knowledge bases.
We develop a lightweight front-end tool for facilitating interactive entity population.
We implement key components necessary for effective interactive entity population: 1) GUI-based dashboards to quickly modify an entity dictionary, and 2) entity highlighting on documents for quickly viewing the current progress.
We aim to reduce user cost from beginning to end, including package installation and maintenance.
The implementation enables users to use this tool on their web browsers without any additional packages --- users can focus on their missions to create entity dictionaries. Moreover, an entity expansion module is implemented as external APIs. This design makes it easy to continuously improve interactive entity population pipelines. We are making our demo publicly available (\url{http://bit.ly/luwak-demo}).
\end{abstract}

\section{Introduction}
Entity extraction is one of the most major NLP components.
Most NLP tools (e.g., NLTK, Stanford CoreNLP, etc.), including commercial services (e.g., Google Cloud API, Alchemy API, etc.), provide entity extraction functions to recognize named entities (e.g., PERSON, LOCATION, ORGANIZATION, etc.) from texts. Some studies have defined fine-grained entity types and developed extraction methods~\cite{Ling2012} based on these types.
However, these methods cannot comprehensively cover domain-specific entities.
For instance, a real estate search engine needs housing equipment names to index these terms for providing fine-grained search conditions. 
There is a significant demand for constructing user-specific entity dictionaries, such as the case of cuisine and ingredient names for restaurant services.
A straightforward solution is to prepare a set of these entity names as a domain-specific dictionary. Therefore, this paper focuses on the {\it entity population} task, which is a task of collecting entities that belong to an entity type required by a user.

We develop LUWAK, a lightweight tool for effective interactive entity population.
The key features are four-fold:
\begin{itemize}
  \item Pure JavaScript implementation to reduce user cost of package installation and maintenance
  \item An entity table dashboard for quickly viewing and modifying a dictionary
  \item A feedback table dashboard for supporting an effective interactive entity population
  \item Entity highlighting on documents for quickly viewing the performance of the current entity dictionary
\end{itemize}
We think these features are key components for effective interactive entity population. 

We choose an interactive user feedback strategy for entity population for LUWAK. A major approach to entity population is {\it bootstrapping}, which uses several entities that have been prepared as a seed set for finding new entities. Then, these new entities are integrated into the initial seed set to create a new seed set. The bootstrapping approach usually repeats the procedure until it has collected a sufficient number of entities. The framework cannot prevent the incorporation of incorrect entities that do not belong to the entity type unless user interaction between iterations. The problem is commonly called {\it semantic drift}~\cite{Curran2007}. Therefore, we consider user interaction, in which feedback is given to expanded candidates, as essential to maintaining the quality of an entity set.
LUWAK implements fundamental functions for entity population, including (a) importing an initial entity set, (b) generating entity candidates, (c) obtaining user feedback, and (d) publishing populated entity dictionary.

We aim to reduce the user’s total workload as a key metric of an entity population tool. That is, an entity population tool should provide the easiest and fastest solution to collecting entities of a particular entity type. User interaction cost is a dominant factor in the entire workload of an interactive tool. Thus, we carefully design the user interface for users to give feedbacks to the tool intuitively. Furthermore, we also consider the {\it end-to-end} user cost reduction.
We adhere to the concept of developing installation-free software to distribute the tool among a wide variety of users, including nontechnical clusters.
This lightweight design of LUWAK might speed up the procedure of the whole interactive entity population workflow. Furthermore, this advantage might be beneficial to continuously improve the whole pipeline of interactive entity population system.

\section{LUWAK: A lightweight tool for interactive entity population}\label{sec:system}
\begin{figure}[t]
\centering
\includegraphics[width=0.45\textwidth]{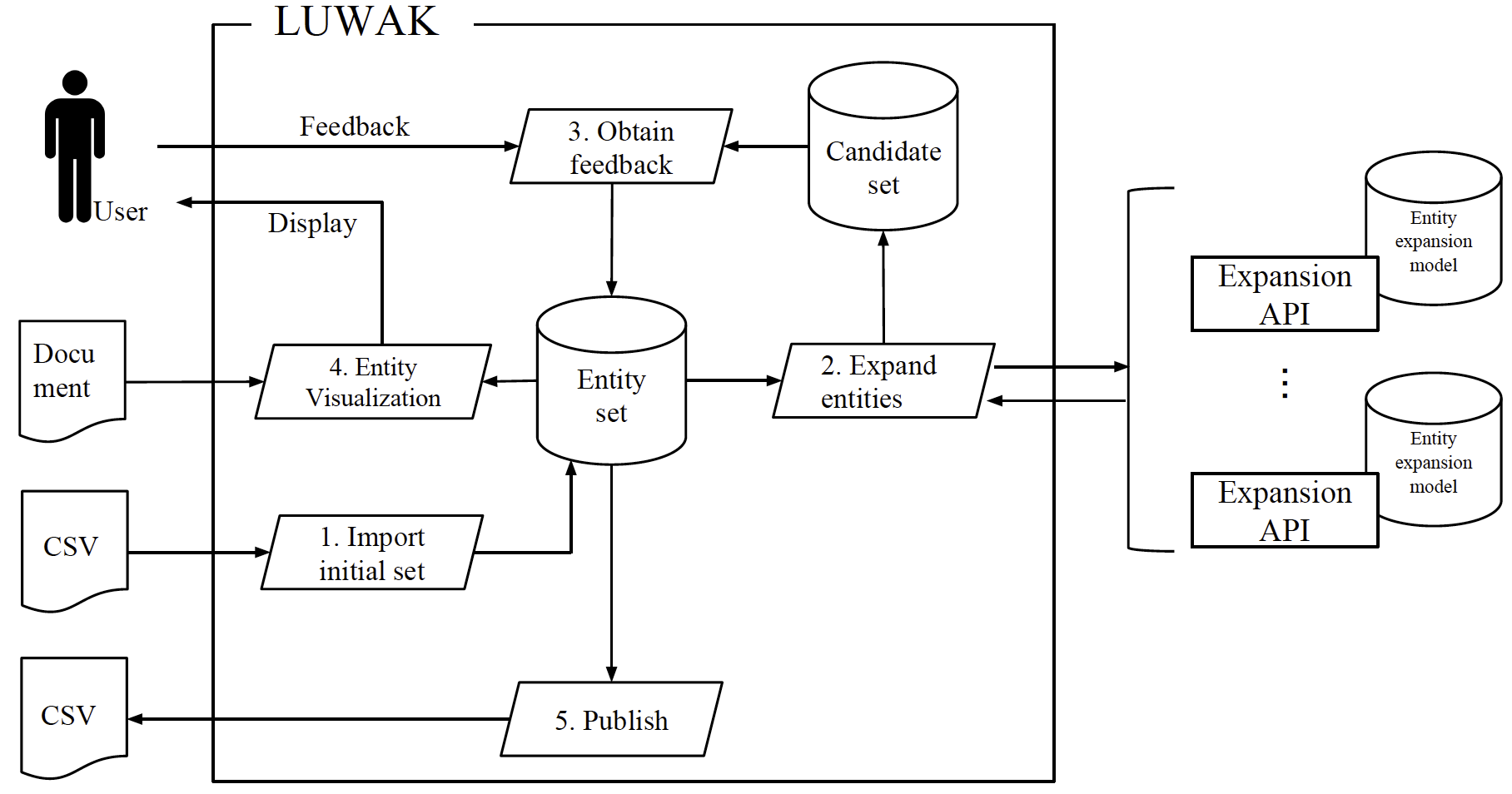}
\caption{Basic workflow of LUWAK.}
\label{fig:workflow}
\end{figure}

Our framework adopts the interactive entity expansion approach. This approach organizes the collaboration of a human worker and entity expansion algorithms to generate a user-specific entity dictionary efficiently. We show the basic workflow of LUWAK in Figure \ref{fig:workflow}. (Step 1) LUWAK assumes that a user prepares an initial seed set manually.
The seed set is shown in the Entity table. (Step 2) A user can send entities in the Entity table to an Expansion API for obtaining entity candidates.
(Step 3) LUWAK shows the entity candidates in the Candidate table for user interaction. Then, the user checks accept/reject buttons to update the Entity table. After submitting the judgments, LUWAK shows the Entity table again. The user can directly add, edit, or delete entities in the table at any time. (Step 4) the user can also easily see how these entities stored in the Entity table appear in a document. (Step 5) After repeating the same procedure (Steps 2--4) for a sufficient time, the user can publish the Entity table as an output.
 
\subsection{Implementation}
LUWAK is implemented in pure JavaScript code, and it uses the LocalStorage of a web browser. A user does not have to install any packages except for a web browser. The only thing the user must do is to download the LUWAK software and put it in a local directory. We believe the cost of installing the tool will keep away a good amount of potential users. This philosophy follows our empirical feeling of {\it the curse of installing required packages/libraries.} Moreover, LUWAK does not need users to consider the usage and maintenance of these additional packages. That is why we are deeply committed to making LUWAK a pure client-side tool in the off-the-shelf style.

\subsection{LUWAK Dashboard}
\begin{figure*}[t]
\centering
    \centering
    \subfigure[]{
        \includegraphics[scale=0.25]{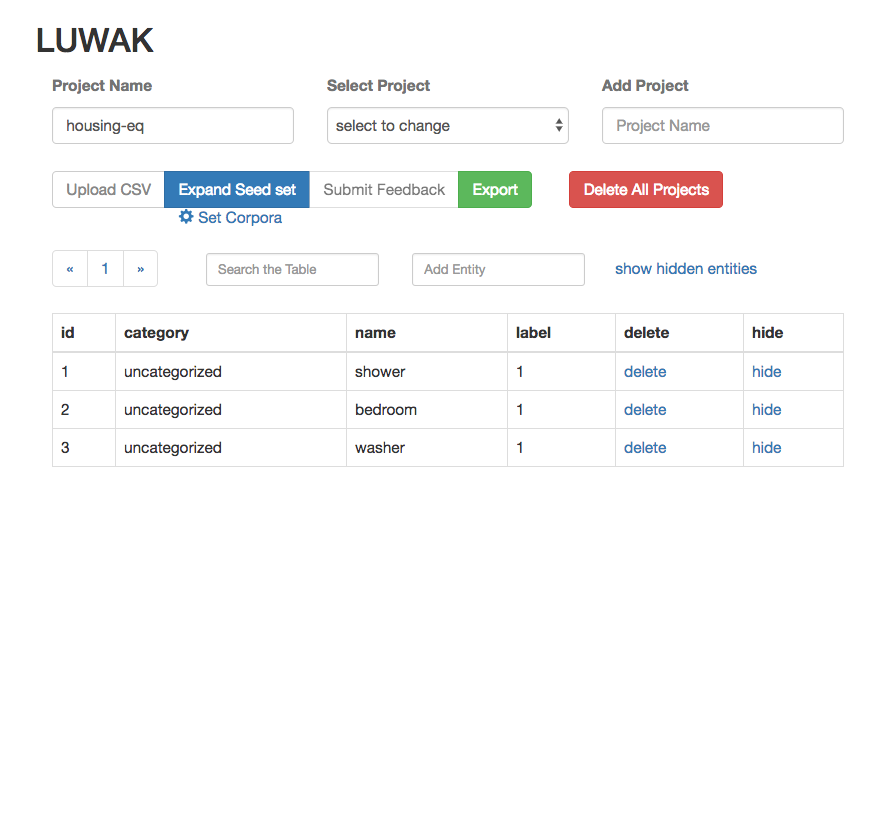} \label{fig:entity_table_screenshot}
    }
    \subfigure[]{
        \includegraphics[scale=0.25]{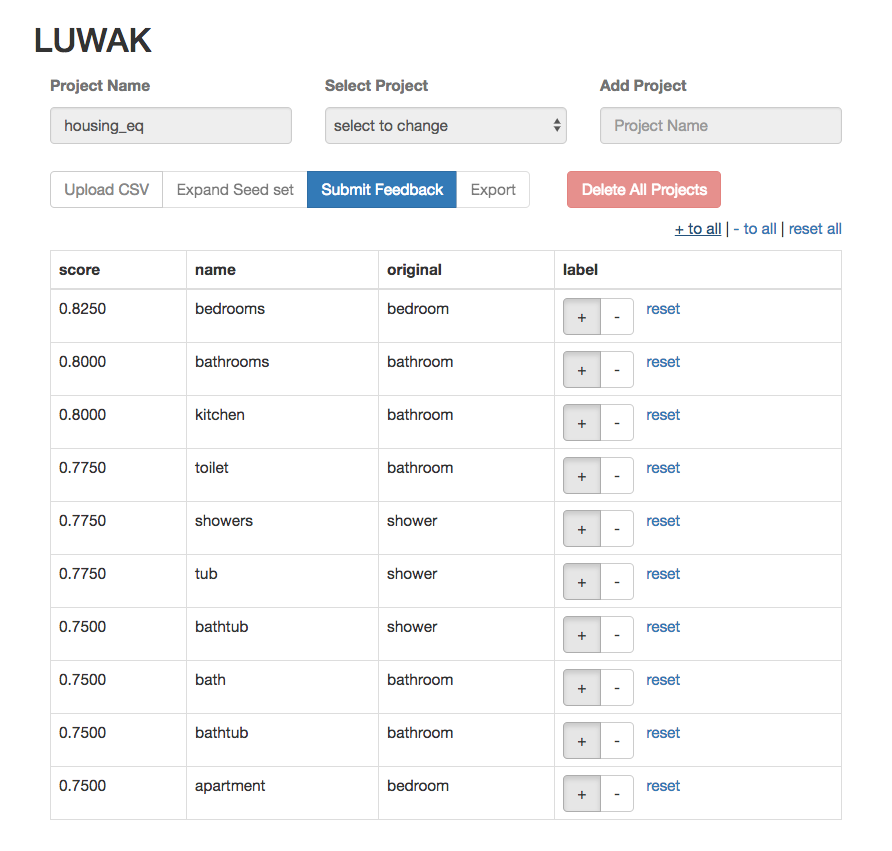} \label{fig:feedback_table_screenshot}
    }
    \subfigure[]{
        \includegraphics[scale=0.25]{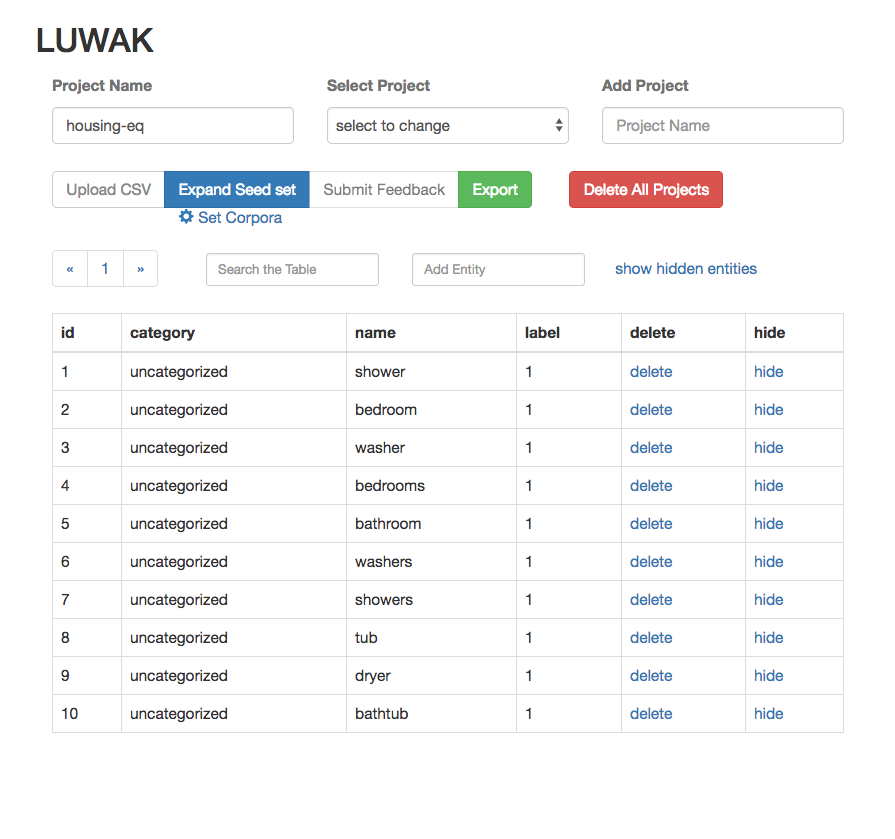} \label{fig:second_entity_table_screenshot}
    }
    \subfigure[]{
        \includegraphics[scale=0.25]{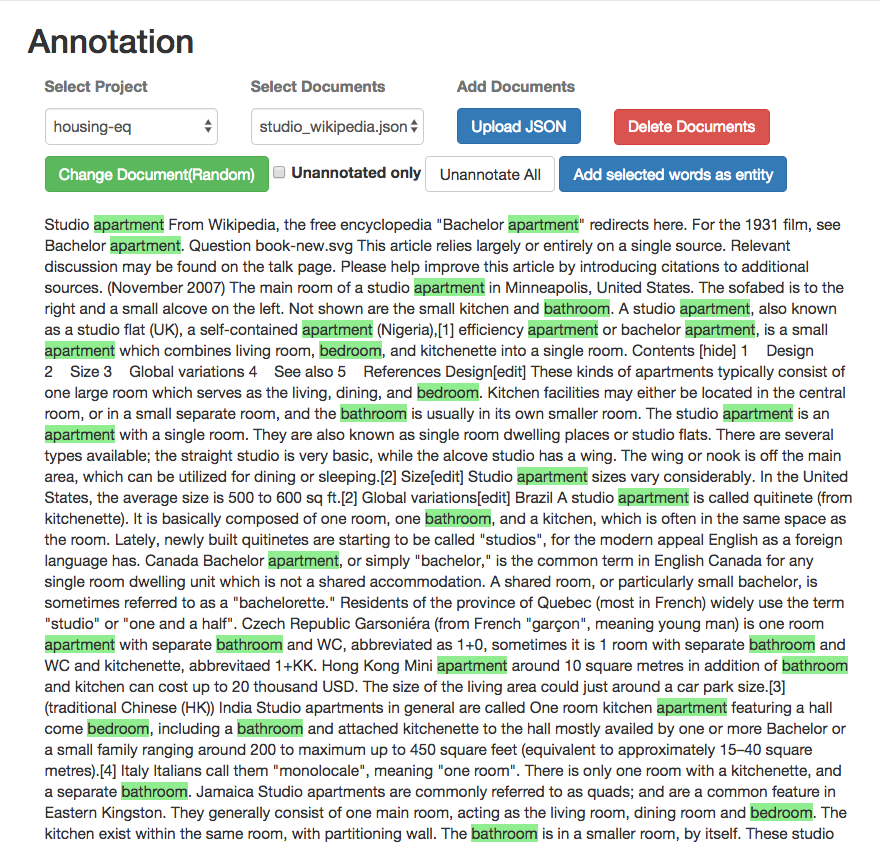} \label{fig:annotation_table_screenshot}
    }

\caption{Screenshots of the LUWAK Dashboard: (a) Entity table shows initial seed entities. (b) Feedback table shows entity candidates generated by the initial seed entities. (c) Entity table has added entities after the submission of generated entity candidates. (d) Entity highlighting function visually emphasizes the words in the current entity set.}
\label{fig:screenshot}
\end{figure*}

LUWAK has a dashboard for quickly viewing an entity dictionary in progress. The dashboard consists of two tables: the Entity table and the Feedback table. The Entity table provides efficient ways to construct and modify an entity dictionary. Figure \ref{fig:entity_table_screenshot} shows the screenshot of the Entity table. The table shows entities in the current entity set. Each row corresponds to an entity entry. Each entry has a {\it label}, which denotes whether the predefined entity type is a positive or a negative example, an {\it original entity}, which was used to find the entity, and the {\it score}, which denotes the confidence score. A user can directly edit the table by adding, renaming, and deleting entities. Moreover, the {\it entity inactivation} function allows a user to manually inactivate entities, so that entity expansion algorithms do not use the inactivated entities. The table implements a page switching function, a search function, and a sorting function to ensure visibility even when there is a large number of entities in the table.

\subsection{Entity Candidate Generation}
\label{subsec:entity_candidate_generation}
 
We design the entity candidate generation module as an external API (Expansion API). The Expansion API receives a set of entities with positive labels.
The Expansion API returns top-$k$ entity candidates.
 
As an initial implementation, we used GloVe~\cite{Pennington2014} as word embedding models for implementing an Expansion API. This API calculates the cosine similarity between a set of positive entities and entities candidates to generate a ranked list.
We prepared models trained based on the CommonCrawl corpus and the Twitter corpus\footnote{\url{http://nlp.stanford.edu/projects/glove/}}. Note that the specification of the expansion algorithm is not limited to the algorithm described in this paper, as LUWAK considers the Expansion API as an external function.

Moreover, we also utilize the category-based expansion module, in which we used is-a relationship between the ontological category and each entity and expanded seeds via category-level. For example, if most of the entities already inserted in the dictionary share the same category, such as Programming Languages, the system suggests that "Programming Language" entities should be inserted in the dictionary when we develop a job skill name dictionary. Category-based entity expansion is helpful to avoid the candidate entity one by one. We used Yago~\cite{Hoffart2013} as an existing knowledge base.




\noindent {\bf External API.} 
In our design of LUWAK, Expansion APIs are placed as an external function outside LUWAK. There are three reasons why we adopt this design. First, we want LUWAK to remain a corpus-free tool. Users do not have to download any corpora or models to start using LUWAK, and it takes too much time to launch an Expansion API server. Second, LUWAK’s design allows external contributors to build their own expansion APIs that are compatible with LUWAK’s interface. We developed the initial version of the LUWAK package to contain an entity Expansion API so users can launch their expansion APIs internally. Third, the separation between LUWAK and the Expansion APIs enables Expansion APIs to use predetermined options for algorithms, including non-embedding-based methods (e.g., pattern-based methods). We can use more than one entity expansion model to find related entities. For instance, general embedding models, such as those built on Wikipedia, might be a good choice in early iterations, whereas more domain-specific models trained on domain-specific corpora might be helpful in later iterations. LUWAK is flexible to change and use more than one Expansion API.
This design encourages us to continuously refine the entity expansion module easily.


\subsection{Example: Housing Equipment Entity Population}
We show an example of populating house equipment entities using LUWAK for improving a real estate search engine. The preliminary step is to prepare seed entities that belong to the real estate house equipment entity type (e.g., kitchen, bath). In this case, a user is supposed to provide several entities ($\sim$ 10) as an initial set of the category. LUWAK first asks the user to upload an initial seed set. The user can add, rename, and delete entities on the Entity table as he or she wants. The user can also choose a set of entity expansion models at any time. Figure \ref{fig:screenshot} shows the entity dashboard in this example.
 
When the user submits the current entity set by clicking the Expand Seed Set button (Figure \ref{fig:entity_table_screenshot}), LUWAK sends a request to the external Expansion APIs that are selected to obtain expanded entities. The returned values will be stored in the Feedback table, as Figure \ref{fig:feedback_table_screenshot} shows. The Feedback table provides a function to capture user feedback intuitively. The user can click the + or - buttons to assign positive or negative labels to the entity candidates. The score column stores the similarity score, which is calculated by the Expansion API as reference information for users. The user can also see how these entities are generated by looking at the original entities in the original column. The original entity information can be used to detect semantic drift. For instance, if the user finds the original entity of some entity candidates has negative labels, the user might consider inactivating the entity to prevent semantic drift.
 
In the next step, the user reflects the feedback by clicking the Submit Feedback button. Then, the user will see the entity dashboard with the newly added entities as shown in Figure \ref{fig:second_entity_table_screenshot}. The user can inactivate the entity by clicking the inactivate button. The user can sort rows by column values to take a brief look at the current entity set. Also, the entity dashboard provides a search function to find an entity for action. The user can also check how entities appear in a test document. As shown in Figure \ref{fig:annotation_table_screenshot}, LUWAK highlights these entities in the current entity set.  After the user is satisfied with the amount of the current entity set in the table, the Export button allows the user to download the entire table, including inactivated entities.

\section{Related Work and Discussion}
Entity population is one of the important practical problems in NLP. Generated entity dictionaries can be used in various applications, including search engines, named entity extraction, and entity linking. Iterative seed expansion is known to be an efficient approach to construct user-specific entity dictionaries. Previous studies have aimed to construct a high-quality entity dictionary from a small number of seed entities~\cite{Ghahramani2005,He2011,Tao2015,Rong2016}.
As we stated in \ref{subsec:entity_candidate_generation}, LUWAK is flexible with the types of algorithms used for entity population. A user can select any combinations of different methods once the Expansion API of the methods are available.
 
Stanford Pattern-based Information Extraction and Diagnostics (SPIED)~\cite{Gupta2014spied} is a pattern-based entity population system.
SPIED requires not only an initial seed set but also document collection because it uses the pattern-based approach. After a user inputs initial seed entities, SPIED generates regular expression patterns to find entity candidates from a given document collection.
This approach incurs a huge computational cost for calculating the scores of every regular expression pattern and every entity candidate in each iteration. Furthermore, SPIED adopts a bootstrapping approach, which does not involve user feedback for each iteration. This approach can easily result in semantic drift.

Interactive Knowledge Extraction~\cite{Dalvi2016} (IKE) is an interactive bootstrapping tool for collecting relation-extraction patterns. IKE also provides a search-based entity extraction function and an embedding-based entity expansion function for entity population. A user can interactively add entity candidates generated by an embedding-based algorithm to an entity dictionary. LUWAK is a more lightweight tool than IKE, which only focuses on the entity population task. LUWAK has numerous features, such as the multiple entity expansion model choices, that are not implemented in IKE. Moreover, LUWAK is a corpus-free tool that does not require a document collection for entity population. Thus, we differentiate LUWAK from IKE, considering it a more lightweight entity population tool.


\section{Summary}
This paper has presented LUWAK, a lightweight front-end tool for interactive entity population. LUWAK provides a set of basic functions such as entity expansion and user feedback assignment. We have implemented LUWAK in pure JavaScript with LocalStorage to make it an installation-free tool. We believe that LUWAK plays an important role in delivering the values of existing entity expansion techniques to potential users including nontechnical people without supposing a large amount of human cost. Moreover, we believe that this design makes it easy to compare performances between interactive entity population pipelines and develop more sophisticated ones.


\begin{small}
\bibliography{luwak}
\bibliographystyle{icml2017}
\end{small}

\end{document}